\documentclass[a4paper]{article}

\usepackage[english]{babel}
\usepackage[utf8]{inputenc}
\usepackage[T1]{fontenc}

\usepackage[a4paper,top=2cm,bottom=2cm,left=3cm,right=3cm,marginparwidth=1.75cm]{geometry}

\usepackage{amsmath}
\usepackage{graphicx}
\usepackage{url}
\usepackage[colorinlistoftodos]{todonotes}
\usepackage[colorlinks=true, allcolors=blue]{hyperref}
\usepackage[rightcaption]{sidecap}
\usepackage{float}
\usepackage[affil-it]{authblk}
\usepackage{amsthm}
\usepackage{graphicx}               
\usepackage{amssymb}
\numberwithin{equation}{subsection}
\usepackage{environ}
\usepackage{bm}
\usepackage[ruled,vlined]{algorithm2e}
\usepackage{tcolorbox}

\usepackage[backend=biber,sorting=ynt]{biblatex}

\NewEnviron{myequation}{%
\begin{equation}
\scalebox{1.5}{$\BODY$}
\end{equation}
}

\title{\textbf{On Bi-gram Graph Attributes}}
\author{\textbf{Thomas Konstantinovsky}}
    \affil{\normalsize Department of Computer Science, Holon Institute of Technology;Golomb 52,Israel}
    \affil{thomaskon90@gmail.com}
    \affil{\normalsize ORCiD: 0000-0002-4448-7928}
\author{\textbf{Matan Mizrachi}}
    \affil{\normalsize Department of Computer Science, Holon Institute of Technology;Golomb 52,Israel}
    \affil{matt8ac@gmail.com}
    \affil{\normalsize ORCiD: 0000-0002-9564-0785}

\begin{document}
\maketitle

\begin{abstract}
We propose a new approach to text semantic analysis and general corpus analysis using, as termed in this article, a "bi-gram graph" representation of a corpus. The different attributes derived from graph theory are measured and analyzed as unique insights or against other corpus graphs. We observe a vast domain of tools and algorithms that can be developed on top of the graph representation; creating such a graph proves to be computationally cheap, and much of the heavy lifting is achieved via basic graph calculations. Furthermore, we showcase the different use-cases for the bi-gram graphs and how scalable it proves to be when dealing with large datasets.
\end{abstract}

\providecommand{\keywords}[1]
{
  \small	
  \textbf{\textit{Keywords---}} #1
}

\section{Introduction}

Corpus representation is central to natural language processing. However, the default approach of representing a corpus usually revolves around the bag of words representation or other vectorization methods. This paper highlights the benefits and use cases of a representation based on inner word relationships derived from the bi-grams of a given corpus.
Previous works that used similar methods revolve around solving a specific problem using a graph representation,\cite{Mass2008HowIM} suggested a new way for grounding the meanings of certain words in sensorimotor categories,\cite{196128} proposed a model of knowledge-based text condensation that resembles today's well-known knowledge-graphs.
\cite{Ngramgraphs} showed promising effectiveness and suggests that n-gram graphs and operators can constitute an effective and updatable text representation method that strengthens some of the algorithms proposed in this article.
However, many graph attributes are left untouched in natural language processing due to the different representations available.
However, many graph attributes are left untouched in natural language processing due to the different representations available.
Previous works also highlighted the benefits of N-gram flexibility with the well-structured representation of directed graphs and their applications towards classification problems\cite{10.3389/fams.2018.00041}.

Representing a corpus as a weighted directed graph of its bi-grams as defined in section 3 not only is computationally cheaper and more scalable in comparison to standard representation methods but also allows a vast domain of analysis as each bi-gram graph represents not only the inner connection between the words and the classic conditional probabilistic model but also a broader semantic meaning that is unique to each corpus. Moreover, as we present in the article, there are novel tools and algorithms that can be applied to those graphs both individually and against different pairs deriving new insights yet to be explored.
The additional benefits gained from the algorithms based on such a representation are similar to the benefits derived via transfer learning when comparing to deep learning methodologies \cite{thrun2012learning}, i.e., after creating a graph that represents the semantics of a particular corpus, the same graph can be used to apply the graph-based algorithms proposed in this article on other corpora in a way using one semantic field to derive information regarding another.

\section{The Bi-gram Graph Representation}

Throughout the entire paper, $BG_x = \left(V, E, W\right)$ is a weighted directed, finite graph, based on corpus 
\begin{center}
    $x = \{t_1, t_2, \dots, t_n : t_i = w_1w_2\dots w_m$ is the $i$th text of the corpus \}
\end{center}with $V$ as the set of words that appears in at least one of $x$'s texts and $E$, which is defined by: \begin{center}
    $E = \{\left(w_p, w_q\right): \left(w_p, w_q\right) \subseteq \bigcup\limits_{i=1}^{n} t_{i} \}$
\end{center} 
The weights set $W$ is defined by:
\begin{center}
    $W = \{\omega\left(e\right) = $ the number of times that $e$ appears as a bi-gram in $x$, $\forall e \in E$ $\}$
\end{center}
Note that punctuation marks has been removed from $x$ in a prior to generating its Bi-gram Graph.
\\
For example, \\ let $x = \{$"I love eating pizza", "I usually enjoy having a pizza when it rains outside",  "The art of making a pizza" $\}$, so its bi-gram graph is the following:

\begin{figure}[H]
\caption{The Bi-gram graph of the corpus $x$}
\centering
\includegraphics[width=0.5\textwidth]{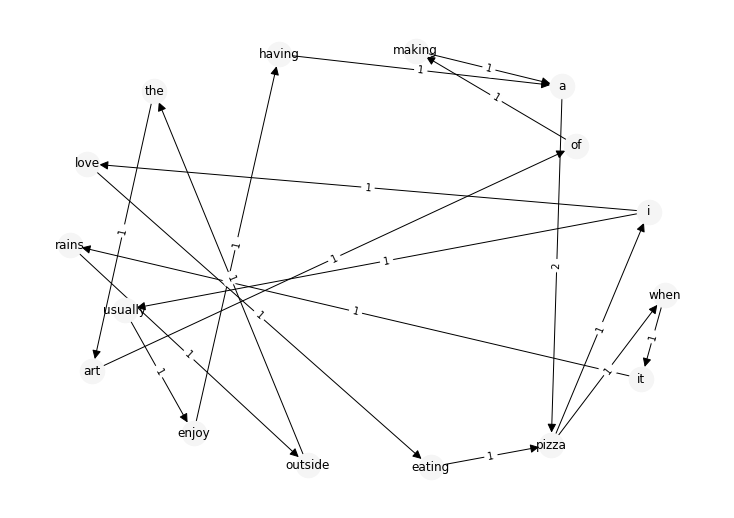}
\end{figure}

\section{Bi-gram Graph Attributes}

\subsection{Chromatic Number}

The main highlight of this article is the different insights gained by deriving the chromatic number and coloring of the graph; mainly, most of the new approaches we present are based on the coloring of the graph and the different relations the graph coloring has within itself and in comparison to other bi-gram graphs.
A graph coloring is achieved by labeling each node with a color such that no two vertices sharing the same edge have the same color.
After coloring a bi-gram graph, each node or each unique word from a given corpus is assigned with a color tag; this tag is heavily dependent on the corpus context as the different edges are formed based on the variability of the sentences in the corpora.
Having a unique context-based color tag for each word by itself provides an interesting meta-feature for different machine learning models, potentially elevating the performance of weak models by introducing context-based features; Moreover, in this article, we propose the "Chromatic Vectorizer," which takes advantage of those tags in order to embed sequences of words based on a bi-gram graph coloring.
As stated in the introduction, the suggested approach of using bi-gram graphs as corpora representation allows for somewhat of a "transfer learning" methodology.
After creating a bi-gram graph for certain corpora and calculating the graph coloring, the same coloring can be used to embed or analyze texts from different corpora by tagging the intersecting words with the chromatic labels of the "pre-trained" bi-gram graph in a way it can be thought of as "context projection" we project the context derived form one corpus on to another.

\subsection{K-Core}

A K-core of graph $BG_x$ is defined as the maximal connected subgraph of $BG_x$ in which all vertices have a degree of at least $k$.
K-cores are mainly mentioned with respect to k-degenerate graphs \cite{lick_white_1970} and have been shown in various cases, such as in \cite{2003498}, that they can be useful when dealing with prediction-related objectives.

We propose a different perspective on the K-core of a graph by addressing problems such as text summarization and dimensionality reduction; the hypothesis we present is that the K-core of $BG_x$ where the maximal $k$ is selected, represents a  summarized version of the text where unlike classical text summarization that focuses on maintaining human readability, the K-core of a bi-gram graph with the maximal $k$ represent the main context of a corpus eliminating any redundant edges that may represent noise or irrelevant data.
From a machine learning perspective, the K-core of a bi-gram graph can be used as an imputer for redundant words by removing all words that have been left out of the K-core; the size of the corpus goes down by a magnitude that is dependent on its original size. 

\subsection{Chromatic Similarity Coefficient}

We propose a new similarity coefficient between two bi-gram graphs based on intersecting nodes and their chromatic numbers, the "chromatic similarity coefficient" - $\Psi(BG_{1},BG_{2})$ defined as:

\theoremstyle{definition}
\newtheorem{definition}{Definition}
\begin{definition}
$$T_{1,2} = \text{The Bi-gram Graph Nodes of Corpora 1 and 2}$$ 
$$ I = \|T_{1} \cap T_{2}\|$$
$$ IC = \|(T_{1} \cap T_{2})  \wedge (T_{1 chrom} \cap T_{2 chrom})\|$$

$$\Psi = \frac{IC}{I}$$
\end{definition}

$\bm{I}$ is defined as the number of nodes in the intersection between the two graphs, i.e., the number of similar unique words in both corpora, and $\bm{IC}$ is defined as the number of intersecting words that also share the same chromatic tag, i.e., the "color" of the node.
An important note to be made is that in the case of an empty intersection, i.e., the two graphs are entirely different; we get a division of zero by zero to avoid such a case $\bm{I}$ is calculated before IC and if $\bm{I} = 0$ then $\bm{IC}$ must also be zero and there is no need to proceed with the formula, formally:

\theoremstyle{definition}
\begin{definition}[$\Psi$ Cases]

$$\Psi(BG_1,BG_2) = \begin{cases}
0 &\text{, $\|I\| = 0 $}\\
1 &\text{, $\|I\|= \|IC\| \ne 0$}
\end{cases}$$

\end{definition}

The proposed similarity coefficient takes advantage of the non-uniform distribution of attributes such as part of speech and named entities across different corpora as shown in section four, e.g., we observed that vast sets of words are usually tagged under the same chromatic label when the context of the corpora is reasonably similar in contrast corpora that are more different yet having the exact words will have a large $\bm{I}$ value but a low $\bm{IC}$ resulting in a low overall $\Psi$ value.

\subsection{Random Chromatic Walker}

The random chromatic walker is yet another approach we propose based on the coloring of the bi-gram graph, and its purpose is to provide a fast and versatile approach to text generation.
The algorithm we propose generates an array that randomly selects a number $J$ such that $1 \leq J \leq \chi(BG_x)$ and using a beta distribution to compensate for the non-uniform distribution of color labels.
The resulting sentence is the concatenation of the paths between two randomly selected words according to the color labels generated in the first array of random color labels.

\begin{algorithm}[H]
\SetAlgoLined
\KwResult{A randomly generated sentence based on a bi-gram graph}
 SNT\_LEN = the length of the randomly generated color label array\;
 CHROM\_VEC = a vector of length equal to SNT\_LEN filled with randomly selected chromatic number as described above, that are smaller or equal to the chromatic number of a given bi-gram graph\;
LAST\_WORD = randomly select a word that is tagged with the color at CHROM\_VEC[0]\;
RESULTING\_SENTENCE = ""\;
 \For{number in CHROM\_VEC[1:]}{
    CUR\_WORD = generate a random word with the color equal to number\;
    PATH = find\_path(LAST\_WORD,CUR\_WORD)\;
    RESULTING\_SENTENCE += PATH[:-1]\;
    LAST\_WORD = CUR\_WORD\;

 }
 \Return RESULTING\_SENTENCE

 \caption{Random Chromatic Walker}
\end{algorithm}

This approach allows for many different and unique results depending on the protocol by which the path between two words was calculated.
In our work, we tested with fair results the following protocols:
\begin{enumerate}
  \item Maximum weight path.
  \item Minimum weight path.
  \item Maximum density path.
  \item Minimum density path.

\end{enumerate}

Where density is defined as the sum of all out and in degrees of a given path, or more formally:$$Density(Path_x) = \sum\nolimits_{\text{firt word in path}}^{\text{last word in path}} { \delta ^-_{word}+\delta ^+_{word}}$$

\subsection{Chromatic Sentence Embedding}

The last graph coloring-based algorithm we propose in this article is a new approach to text embedding. When working with machine learning models, the standard approach is to transform a given text into some vector space representation in order to be able to feed the text into a model; common text vectorization algorithms range from observational methods such as the bag of words \cite{doi:10.1080/00437956.1954.11659520} and TF-IDF vectorization \cite{AIZAWA200345} to more advanced and state of the art deep learning model such as GloVe \cite{inproceedings} and word2vec \cite{mikolov2013efficient}, all these approaches differ in both in their benefits and significance with respect to a given task as well as in their computational complexity.
The "chromatic vectorization" proposed by us performs an embedding computational similar to the bag of words approach, but instead of representing each text as a representation of words frequency, the chromatic vectorizer embeds the context in a given corpus.
The algorithm for chromatic vectorization is dependent on the coloring of a given bi-graph graph, meaning that, similarly to earlier statements, a transfer learning approach can be applied, and a graph of a corpus with the desired context can be used to chromatic vectorizer a given text.
The algorithm is defined as:

\begin{algorithm}[H]
\SetAlgoLined
\KwResult{A text embedding based on a a graph coloring}
 initialization\;
 chromatic\_vector = array of zeros with length equal to the number of words in text\;
 chromatic\_dictionary = a dictionary extracted from a bi-gram graph coloring that maps each node to a color label\; 
 \For{index,word in \textbf{enumerate}(text\_words)}{
 \eIf{word not in chromatic\_dictionary}{
chromatic\_vector[index] = $-1$\;
   }{
  chromatic\_vector[index] = chromatic\_dictionary[word]\;
  }

 }
\Return chromatic\_vector
 \caption{Chromatic Embedding}
\end{algorithm}

An important note to be made as for the chromatic vectorizer concerns the hypothesis of its efficiency; the chromatic embedding space does not meet the requirements of an injective function, meaning two different sentences can have the same chromatic embedding, a question arises when analyzing such property as for the contextual meaning of two sentences that are mapped to the same embedding yet are different, and we yet found such a case after preforming the embedding algorithm on over 20 datasets ranging from 4,000 samples up to 150,000 samples from different contextual themes.
We hypothesize that sentences mapped to the same vector represent the same underlying context with respect to a certain theme observed in a bi-gram graph. 

\section{Results}

Throughout the research, the different methods and metrics proposed in this article were applied to over 20 corpora of different and similar contexts. The results shown below showcase the different outcomes derived using the methods in this paper.
\subsection{Graph Coloring POS and NER Interrelations}

\begin{figure}[H]
  \centering
  \begin{minipage}[H]{0.4\textwidth} 
    \includegraphics[width=\textwidth]{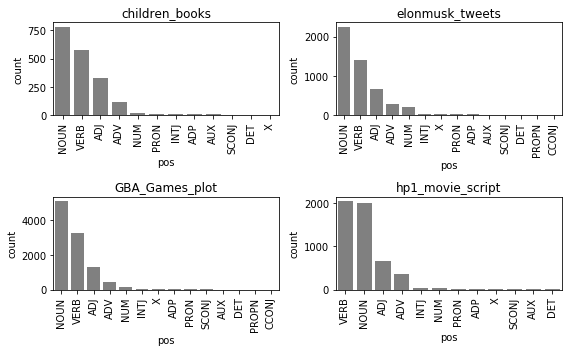}
    \caption{Distribution of POS Per Color Label}
  \end{minipage}
  \hfill
  \begin{minipage}[H]{0.4\textwidth} 
    \includegraphics[width=\textwidth]{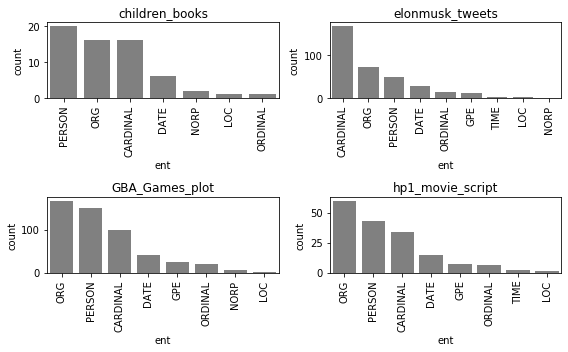}
    \caption{Distribution of NER Per Color Label}
  \end{minipage}
\end{figure}

There is a key note to be made as to the distribution of different attributes associated with color tags; the first is regarding the non-uniform distribution of part-of-speech tags associated with each tag; we observed that in all the corpora we tested, the same distribution has appeared where the first color tag is mostly made out of nouns the second from verbs an so on as can be seen in Figure 1, the hypothesis we propose goes to a new approach of part of speech tagging using the insight derived, i.e., replacing the current computationally expansive part-of-speech algorithms and assigning a part of speech tag according to a given color tag with some deviation as the part of speech distribution in each color tag contains other tags as well.

\subsection{K-core Dimensionality Reduction}
The K-core dimensionality reduction and noise reduction approach we proposed in section 3.2 demonstrated below shows outstanding results when used in classification-based machine learning pipelines.
In the example shown below, a corpus of spam and ham SMSes\footnote{https://www.dt.fee.unicamp.br/~tiago/smsspamcollection/} was taken as a classic example for text classification in machine learning natural language processing; the corpus was cleared of stopwords and converted into a bag of words representation.
Using a naive Random Forest model with default parameters, we received an accuracy score of $97\%$, the confusion matrix of that same process is demonstrated in Figure 4.
The bi-gram graph of the SMS corpus contains $10730$ nodes and $41389$ as we hypothesized earlier; many of these nodes are redundant and potentially can be imputed.
The K-core of maximal K in this graph results in a new graph of 165
nodes and $3736$ edges, using only the words left in the K-core graph, the same naive Random Forest model with default parameters achieved an accuracy of $91\%$, and its confusion matrix can be seen in Figure 5.
After reducing the original quantity of unique words to only $10\%$, we observe only a minor reduction in accuracy.

\begin{figure}[H]
  \centering
  \begin{minipage}[H]{0.4\textwidth} 
    \includegraphics[width=\textwidth]{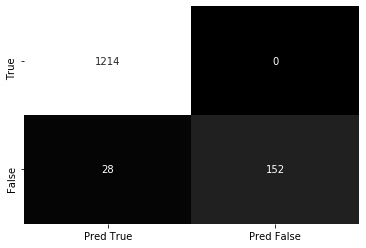}
    \caption{Before Imputing Words Outside of The K-Core}
  \end{minipage}
  \hfill
  \begin{minipage}[H]{0.4\textwidth} 
    \includegraphics[width=\textwidth]{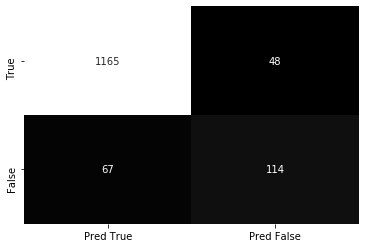}
    \caption{After Imputing Words Outside of The K-Core}
  \end{minipage}
\end{figure}

In comparison to traditional dimensionality reduction algorithms such as principal component analysis or singular value decomposition, the result of the K-core extraction is directly interpretable and are a result of imputation, unlike the formerly mentioned algorithms, which project the original data onto new vector spaces that are much less trivial to interpret with respect to the original data.
Also, note that there is one degree of freedom when using K-core as a method to reduce the dimensionality, although selecting the maximum K of a graph allows extraction of the most significant context and connections between the words in the original corpus, lower K values can be selected retaining more information and more dimensions.

\subsection{Chromatic Similarity Coefficient}
When analyzing the results of the \emph{Chromatic Similarity Coefficient} computation between our corpora,
 \begin{figure}[H]
\caption{$\Psi$ Similarity Matrix of corpora \cite{RestaurantCustomerReviews}, \cite{Anime_synopsis}, \cite{children_books} and \cite{nirvana_lyrics}}
\centering
\includegraphics[width=0.5\textwidth]{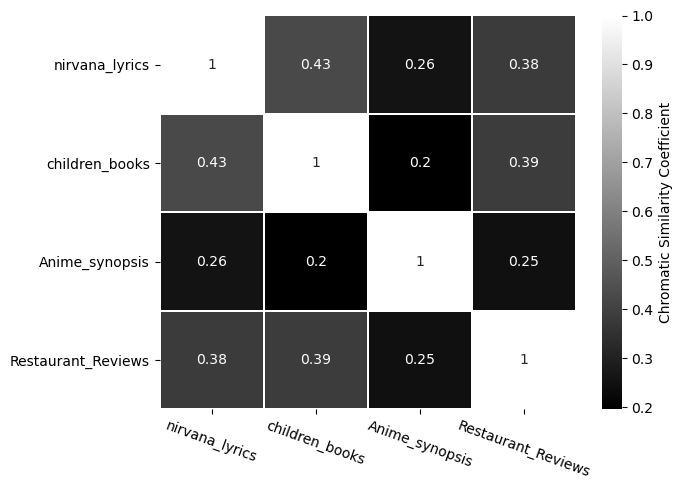}
\end{figure}
 
  we wanted to estimate the quality of the \emph{Chromatic Similarity Coefficient} as a similarity coefficient, in comparison with other similarity measures, such as \emph{Cosine Similarity} and \emph{Jaccard Index}\cite{https://doi.org/10.1111/j.1469-8137.1912.tb05611.x}. 
  The \emph{Cosine Similarity} measure is defined by: \begin{center}
      $\cos\left(A, B\right) = \frac{A\cdot B^T}{\|A\| \cdot \|B\|} : A, B \in \mathbb{R}^n$
  \end{center}
 where, in our case, $A, B$ are the TF-IDF embeddings of corpora $x, y$ respectively,
 
and the \emph{Jaccard Index} measure, is defined by: \begin{center}
    $J\left(A, B\right) = \frac{A \cap B}{A \cup B}$
\end{center}
where $A, B$ are $x, y$ corpora's sets of unique words, respectively.

\begin{figure}[H]
  \centering
  \begin{minipage}[H]{0.4\textwidth} 
    \includegraphics[width=\textwidth]{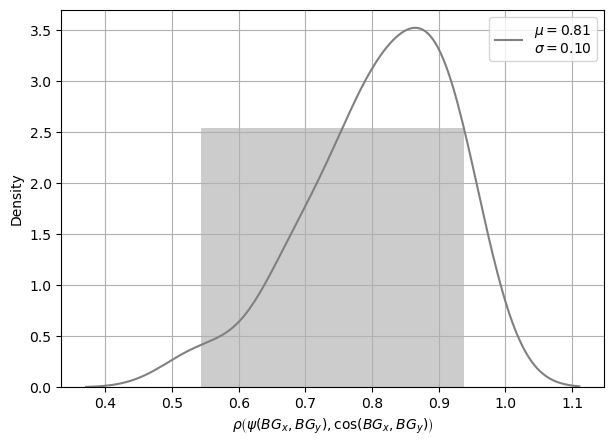}
    \caption{\emph{Cosine Similarity} in Relation to $\Psi$}
  \end{minipage}
  \hfill
  \begin{minipage}[H]{0.4\textwidth} 
    \includegraphics[width=\textwidth]{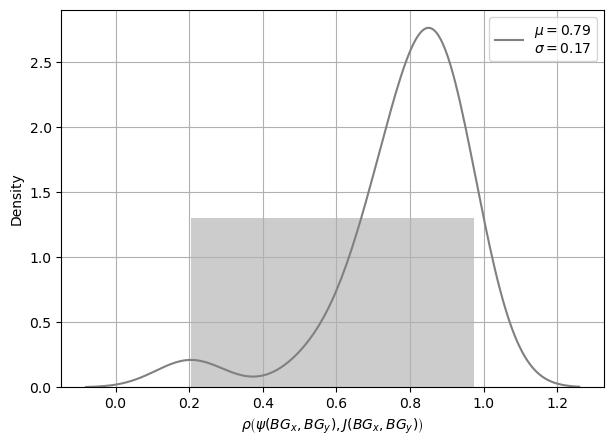}
    \caption{\emph{Jaccard Index} in Relation to $\Psi$}
  \end{minipage}
\end{figure}

Our comparison is done by examining the distribution of the \emph{Pearson's correlation Coefficient}, that is denoted by $\rho$, between each pair of our corpora \emph{Chromatic Similarity Coefficient} to their \emph{Cosine Similarity} and \emph{Jaccard Index}.
The results shown in Figures 7 and 8 demonstrate how $\Psi$ is similar on average to both \emph{Jaccard Index} and \emph{Cosine Similarity} based on TF-IDF embedding, even though the formula of $\Psi$ is closer to the \emph{Jaccard Index} formula, there is a stronger correlation on average between the TF-IDF embedded \emph{Cosine Similarity}.
Overall, even though $\Psi$ is different to the tested similarity measures, it brings an additional contextual consideration that is represent via the coloring of the bi-gram graph i.e. the $IC$ component in $\Psi$'s formula.

\subsection{Chromatic Random Walker}
When testing the chromatic random walker, different path protocols lead to various results depending on the context of the corpus on which the bi-gram graph was built; the below examples show three different sentences generated using three different protocols on three different corpora.

\begin{tcolorbox}
\textbf{Corpus}: Nirvana Lyrics;
\textbf{Protocol}: Heaviest Path
\begin{verbatim}
'calm inside seems think happy think one wanna fun think happy know
right sure killed toe jam booger stomach acid'
\end{verbatim}
\end{tcolorbox}

\begin{tcolorbox}
\textbf{Corpus}: Terminator 3 Script;
\textbf{Protocol}: Lightest Path
\begin{verbatim}
'like i see you dying never we cry to reasons die birds fly happily when we take off'
\end{verbatim}
\end{tcolorbox}

\begin{tcolorbox}
\textbf{Corpus}: Movie Synopses;
\textbf{Protocol}: Heaviest Density
\begin{verbatim}
'rescue young doctor because finding mysterious deaths young police man
gods mean nothing'
\end{verbatim}
\end{tcolorbox}

Note that stop words were removed from all corpora during preprocessing, yet some protocols achieve remarkably good readability and logic results.
One of the most significant advantages of the chromatic random walker is its versatile nature in that it can adapt different scoring protocols for paths and different path traversals.

\section{Future Work}

The main purpose of this paper was to introduce different use-cases and metrics derived from unexplored bi-gram graph attributes, mainly the chromatic number and the coloring of such graphs.
Ongoing and future work includes establishing robust rules and methods concerning the coloring of a bi-gram graph and the different patterns emerging when scaling it to many datasets. We have seen that the coloring of bi-gram graphs allows us to derive a new similarity coefficient and various text generation and dimensionality reduction techniques. However, graphs associated with datasets of a similar topic have a similar structure. We intend to describe how such datasets interrelate, yielding a solution that can unfold a computationally cheap solution for large-scale corpus analysis.

\printbibliography

@article{Mass2008HowIM,
  title={How Is Meaning Grounded in Dictionary Definitions?},
  author={A. Mass{\'e} and G. Chicoisne and Y. Gargouri and S. Harnad and Olivier Picard and Odile Marcotte},
  journal={ArXiv},
  year={2008},
  volume={abs/0806.3710}
}

@INPROCEEDINGS{196128,
  author={Reimer, U. and Hahn, U.},
  booktitle={[1988] Proceedings. The Fourth Conference on Artificial Intelligence Applications}, 
  title={Text condensation as knowledge base abstraction}, 
  year={1988},
  volume={},
  number={},
  pages={338-344},
  doi={10.1109/CAIA.1988.196128}}

@inbook{Ngramgraphs,
author = {Giannakopoulos, George and Karkaletsis, Vangelis},
year = {2009},
month = {01},
pages = {},
title = {N-gram graphs: Representing documents and document sets in summary system evaluation}
}

@ARTICLE{10.3389/fams.2018.00041,
  
AUTHOR={Violos, John and Tserpes, Konstantinos and Varlamis, Iraklis and Varvarigou, Theodora},   
	 
TITLE={Text Classification Using the N-Gram Graph Representation Model Over High Frequency Data Streams},      
	
JOURNAL={Frontiers in Applied Mathematics and Statistics},      
	
VOLUME={4},      

PAGES={41},     
	
YEAR={2018},      
	  
URL={https://www.frontiersin.org/article/10.3389/fams.2018.00041},       
	
DOI={10.3389/fams.2018.00041},      
	
ISSN={2297-4687},   
}

@book{thrun2012learning,
  title={Learning to Learn},
  author={Thrun, S. and Pratt, L.},
  isbn={9781461555292},
  url={https://books.google.co.il/books?id=X\_jpBwAAQBAJ},
  year={2012},
  publisher={Springer US}
}

@article{lick_white_1970, title={k-Degenerate Graphs}, volume={22}, DOI={10.4153/CJM-1970-125-1}, number={5}, journal={Canadian Journal of Mathematics}, publisher={Cambridge University Press}, author={Lick, Don R. and White, Arthur T.}, year={1970}, pages={1082–1096}}

@article{2003498,
  title={Prediction of Protein Functions Based on K-Cores of Protein-Protein Interaction Networks and Amino Acid Sequences},
  author={Md. Altaf-Ul-Amine and Kensaku Nishikata and Toshihiro Korna and Teppei Miyasato and Yoko Shinbo and Md. Arifuzzaman and Chieko Wada and Maki Maeda and Taku Oshima and Hirotada Mori and Shigehiko Kanaya},
  journal={Genome Informatics},
  volume={14},
  number={ },
  pages={498-499},
  year={2003},
  doi={10.11234/gi1990.14.498}
}

@article{doi:10.1080/00437956.1954.11659520,
author = {Zellig S. Harris},
title = {Distributional Structure},
journal = {\textit{WORD}},
volume = {10},
number = {2-3},
pages = {146-162},
year  = {1954},
publisher = {Routledge},
doi = {10.1080/00437956.1954.11659520},

URL = { 
        https://doi.org/10.1080/00437956.1954.11659520
    
},
eprint = { 
        https://doi.org/10.1080/00437956.1954.11659520
    
}

}

@article{AIZAWA200345,
title = {An information-theoretic perspective of tf–idf measures},
journal = {Information Processing and Management},
volume = {39},
number = {1},
pages = {45-65},
year = {2003},
issn = {0306-4573},
doi = {https://doi.org/10.1016/S0306-4573(02)00021-3},
url = {https://www.sciencedirect.com/science/article/pii/S0306457302000213},
author = {Akiko Aizawa},
}

@inproceedings{inproceedings,
author = {Pennington, Jeffrey and Socher, Richard and Manning, Christopher},
year = {2014},
month = {01},
pages = {1532-1543},
title = {Glove: Global Vectors for Word Representation},
volume = {14},
journal = {EMNLP},
doi = {10.3115/v1/D14-1162}
}

@misc{mikolov2013efficient,
      title={Efficient Estimation of Word Representations in Vector Space}, 
      author={Tomas Mikolov and Kai Chen and Greg Corrado and Jeffrey Dean},
      year={2013},
      eprint={1301.3781},
      archivePrefix={arXiv},
      primaryClass={cs.CL}
}

@misc{RestaurantCustomerReviews,
    title={Restaurant Customer Reviews}, 
    author={@Vicky, Kaggle}, 
    year={2019},
    url={https://www.kaggle.com/vigneshwarsofficial/reviews}}

@misc{nirvana_lyrics,
    title={Nirvana Lyrics}, 
    author={@darkrubiks, Kaggle}, 
    year={2021},
    url={https://www.kaggle.com/darkrubiks/nirvana-lyrics}}

@misc{children_books,
    title={Highly Rated Children Books And Stories}, 
    author={@Thomas Konstantin, Kaggle}, 
    year={2021},
    url={https://www.kaggle.com/thomaskonstantin/highly-rated-children-books-and-stories}}

@misc{Anime_synopsis,
    title={Top 10000 Anime Movies ,OVA's and Tv-Shows}, 
    author={@Thomas Konstantin, Kaggle}, 
    year={2021},
    url={https://www.kaggle.com/thomaskonstantin/top-10000-anime-movies-ovas-and-tvshows}}

@article{https://doi.org/10.1111/j.1469-8137.1912.tb05611.x,
author = {Jaccard, Paul},
title = {THE DISTRIBUTION OF THE FLORA IN THE ALPINE ZONE.1},
journal = {New Phytologist},
volume = {11},
number = {2},
pages = {37-50},
doi = {https://doi.org/10.1111/j.1469-8137.1912.tb05611.x},
url = {https://nph.onlinelibrary.wiley.com/doi/abs/10.1111/j.1469-8137.1912.tb05611.x},
eprint = {https://nph.onlinelibrary.wiley.com/doi/pdf/10.1111/j.1469-8137.1912.tb05611.x},
year = {1912}
}

\end{document}